%% file: main.tex
\pdfoutput=1

\documentclass[11pt]{article}

\usepackage{acl}

\usepackage{times}
\usepackage{latexsym}

\usepackage[T1]{fontenc}

\usepackage[utf8]{inputenc}

\usepackage{microtype}

\definecolor{green_plt}{HTML}{2CA02C}

\usepackage{standalone}
\usepackage[nointegrals]{wasysym} %
\usepackage{caption}
\usepackage{subcaption}
\usepackage{graphicx,float}
\usepackage{booktabs,multirow}
\usepackage{amsmath}

\title{Instructions for *ACL Proceedings}

\begin{document}

\author{Oskar van der Wal, Jaap Jumelet, Katrin Schulz, Willem Zuidema \\
    Institute for Logic, Language and Computation, University of Amsterdam \\
    \texttt{\{o.d.vanderwal,j.w.d.jumelet,k.schulz,w.h.zuidema\}@uva.nl} \\}

\title{The Birth of Bias: A case study on the evolution of gender bias in an English language model}

\maketitle

\input{sections/abstract}

\input{sections/introduction}

\input{sections/method}

\input{sections/experiment1}

\input{sections/experiment2}

\input{sections/experiment3}

\input{sections/discussion}

\input{sections/conclusion}

\section*{Acknowledgements}
This publication is part of the project “The biased reality of online media - Using stereotypes to make media manipulation visible” (with project number 406.DI.19.059) of the research programme Open Competition Digitalisation-SSH, which is financed by the Dutch Research Council (NWO).

We want to thank Maartje~ter~Hoeve for her feedback on an earlier version of the paper.

\bibliography{references,InterpretabilityBiasinNLP}
\bibliographystyle{acl_natbib}

\clearpage
\input{sections/appendix}

\end{document}

%% file: sections/abstract.tex
\begin{abstract}
Detecting and mitigating harmful biases in modern language models are widely recognized as crucial, open problems. 
In this paper, we take a step back and investigate how language models come to be biased in the first place.
We use a relatively small language model, using the LSTM architecture trained on an English Wikipedia corpus. With full access to the data and to the model parameters as they change during every step while training, we can map in detail how the representation of gender develops, what patterns in the dataset drive this, and how the model's internal state relates to the bias in a downstream task (semantic textual similarity).
We find that the representation of gender is dynamic and identify different phases during training.
Furthermore, we show that gender information is represented increasingly locally in the input embeddings of the model and that, as a consequence, debiasing these can be  %
effective in reducing the downstream bias.
Monitoring the training dynamics, allows us to detect an asymmetry in how the female and male gender are represented in the input embeddings. This is important, as it may cause naive mitigation strategies to introduce new undesirable biases.
We discuss the relevance of the findings for mitigation strategies more generally and the prospects of generalizing our methods to larger language models, the Transformer architecture, other languages and other undesirable biases. 
\end{abstract}

%% file: sections/introduction.tex
\section{Introduction}

Large Language Models (LLMs), such as BERT \cite{tenney_bert_2019} and GPT-3 \cite{brownLanguageModelsAre2020}, have become crucial building blocks of many AI systems \citep{bommasani_opportunities_2021}. As these models are used in ever more real world applications, it has become increasingly important to monitor, understand and mitigate the harmful behaviours they may exhibit. 
In particular, many of those LLMs have been shown to learn undesirable biases towards certain social groups \cite{bender2021stochasticparrots,weidinger2021ethical}. 
These biases pose a serious threat for the usefulness of the technology, as they may unfairly influence the decisions, recommendations or texts that AI systems building on those LLMs generate. If we want to keep exploring the immense potential of the technology, we need to find ways to avoid or at least mitigate unwanted biases in language models. 

However, detecting, mitigating and even defining undesirable biases have proven to be extremely challenging tasks. %
One key difficulty is deciding on where in the language modelling pipeline to measure and to intervene: in the data used for training, in the internal representations of the models, or only in the applications that are built on top of the language models (the \emph{downstream applications})? 
Many recent papers have proposed methods that work at one or two of these loci, 
for example, by focusing on the dataset \cite{dixon2018MeasuringMitigatingUnintended,hallmaudslay2019ItAllName,lu2020GenderBiasNeural}, %
the training procedure \cite{zhang2018mitigating,zhaoLearningGenderNeutralWord2018,liu2019incorporating}, 
or on measuring and fixing biases in word embeddings or internal states of language models \cite{bolukbasi2016ManComputerProgrammer,ethayarajhUnderstandingUndesirableWord2019,wang2020DoubleHardDebiasTailoring,basta2019EvaluatingUnderlyingGender,may2019MeasuringSocialBiases,kurita2019MeasuringBiasContextualized,tan_reliability_2021}.

In this paper, we do not choose one of these loci, but rather aim to reach an understanding of how they all three relate to each other: how do patterns in the dataset yield a particular structure in the internal states of the language model, and how does this internal structure, in turn, lead to  biased behaviour in a downstream task?
To answer these difficult questions, %
we constrain our work quite radically. First, we work with an LSTM language model and dataset that, although still involving $\sim$90 million words, is small compared to some recent, high-profile LLMs. By doing so we have full control of the training of the model, and full access to the dataset and the internal states of the model at many intermediate points (\emph{checkpoints}) during training. Second, we limit ourselves to only a single, heavily studied bias: gender bias (measured along a female-to-male gender axis) in English
. 
This allows us to make use of many tools already developed for this task, including measures for bias applicable to each of the components of the language modelling pipeline %
and a method for debiasing. 

With this setup, we study how strongly bias measurements in the various stages of the pipeline correlate, how the representations and correlations evolve over training time, and establish a causal link between the identified representation of gender and downstream bias. This provides a uniquely detailed view on the birth of one type of bias, as well as some key lessons that we expect to be useful for detecting and mitigating other biases, in other language models and other languages as well.

%% file: sections/method.tex
\section{Approach}
In our experiments, we study the evolution of gender bias in different representations of an English LSTM language model.\footnote{Our code can be found at \url{https://github.com/bias-barometer/birth-of-bias}.}
We explain how we define gender bias in this particular context and motivate our approach in relation to understanding the source of downstream representational harms, but how we operationalize gender bias is explained in Sections \ref{sec:experiments-gender-representation} and \ref{sec:experiments-characterizing} when discussing the experiments.

\paragraph{Gender bias}
We understand bias as a systematic deviation in behaviour from a norm.
As our focus is on gender bias in language models, the relevant behaviour we are measuring is how strongly certain words or concepts (in our case occupation terms such as \emph{nurse} or \emph{carpenter}) are associated by the model with one gender instead of another. This strength of association can be measured in different ways and at different points in the language modelling pipeline. In particular, we will look at bias in  internal representations of the model and in its output behaviour. Ideally, the strength of association should be equal for different genders. If the model deviates from this norm, we say that the model exhibits gender bias.\footnote{Because we heavily rely on existing tools from previous works for measuring gender bias, we restrict ourselves to representing gender along a female-to-male axis. We recognize that this is an unfortunate simplification \cite[e.g.][]{west1987doing,richards2016non} and hope to overcome this limitation in future work.}

Whether bias in a language model causes harm, depends on the downstream application of the model and what constitutes fair and just behaviour in this particular context, but we believe that a detailed understanding of how bias is learned by and represented in these models can facilitate the development of methods to counteract bias that are tailored to a particular application and the potential harm bias can cause in that context.

With this broad scope, we hope to learn how the language model represents gender stereotypes of occupations in earlier representations of the pipeline, and how these may help explain representational harms~\cite{blodgett2020language} downstream. For instance, if a language model with gender stereotypes for occupations is used in a translation system, it may propagate the undesirable world-view of all doctors being male and all nurses being female.
Understanding how these stereotypical representations come about can help in developing new detection and mitigation strategies for these and other stereotypes in AI systems building on language models.

\paragraph{The LSTM language model}
\label{sec:language_model}

In this paper, we study the gender bias of an LSTM language model \citep{hochreiter97}. 
We follow the setup from \citet{gulordava2018colorless}, and train the model on their training set of \textasciitilde 90M tokens, with a vocabulary of 50,000 (full-word) tokens, extracted from the English Wikipedia corpus. %
Following \citeauthor{gulordava2018colorless}, we lower the learning rate at epoch 20 using a plateau scheduler.
Our training regime differs in one aspect: we use weight-tying for the encoder and decoder \citep{press-wolf-2017-using}.
We make this adjustment to simplify our analysis, as it leads to a smaller model size with comparable performance and limits the available static word vectors to one embedding space instead of two.

We train three language models with different random seeds for 40 epochs, where an epoch is defined as one full pass through all the training data.
During training, we save intermediate checkpoints of the LSTM in order to examine how its behaviour develops over time.
Because model behaviour changes most drastically in the first epoch, we save checkpoints with a higher granularity for that phase.
\\\\
In the rest of this paper, we investigate the representation of gender (bias) in three components of the language modelling pipeline: (i) the dataset, (ii) the input embeddings (provided by the encoder), and (iii) the downstream behaviour (a semantic textual similarity task).

%% file: sections/experiment1.tex
\section{The evolution of gender representation in the input embeddings}
\label{sec:experiments-gender-representation}

In order for a model to acquire undesirable gender biases, it first needs to build up a representation of the concept of gender. In fact, gender bias can be seen as an extension of this concept to words to which we do not want to assign gender. Understanding the process of how a model develops a representation of gender is, therefore, an important part of understanding the evolution of gender bias. For this reason, we start in this section with an investigation of the learning dynamics of gender in general. 
We will focus our analysis on gender representations within input embeddings. %

\begin{figure}[ht]
    \centering
    \includegraphics[width=\columnwidth]{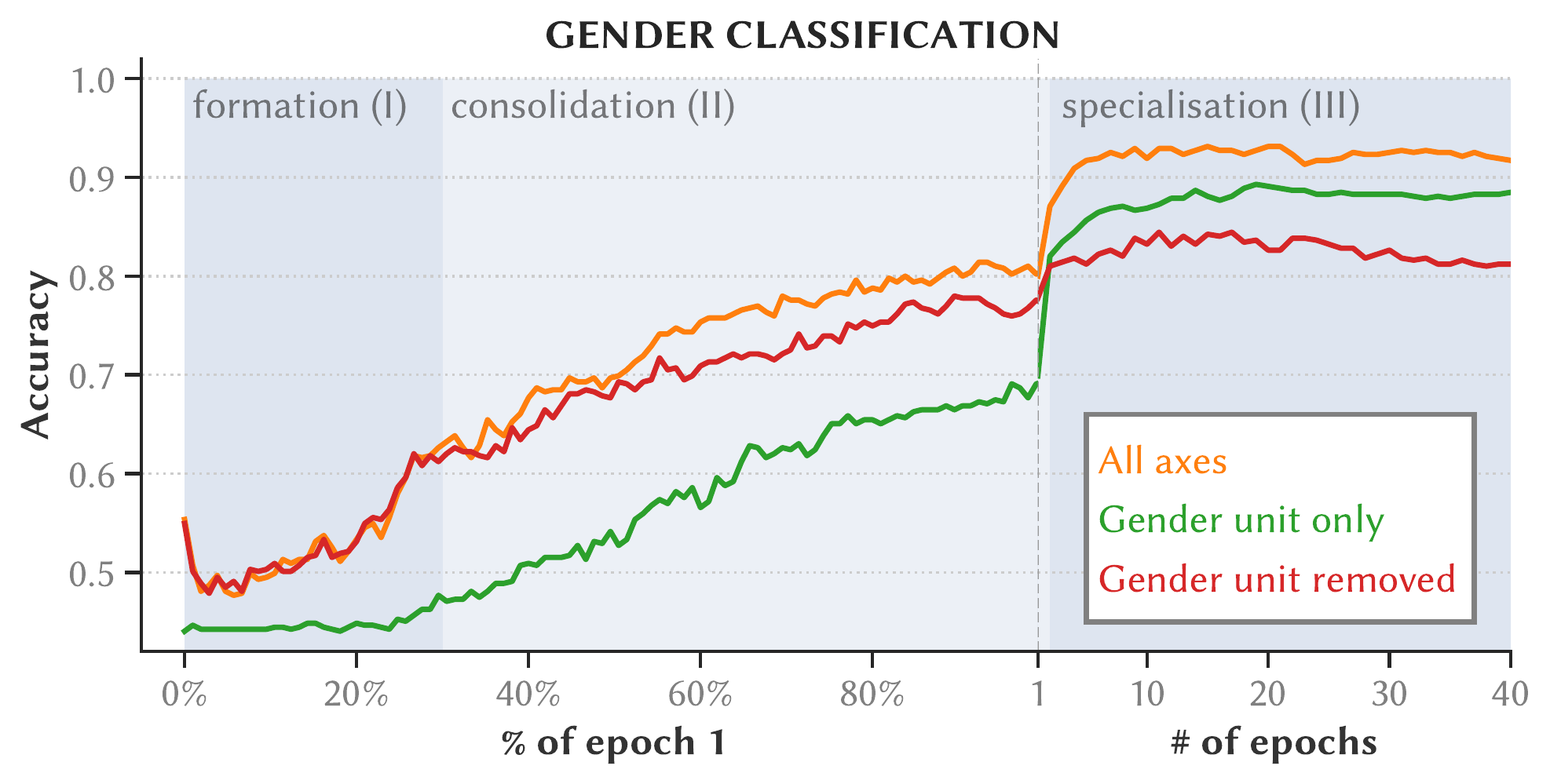}
    
    \caption{Classification accuracy of gender using three different classifiers, that use only the dominant gender unit (green), all other units (red), or all units (orange). Curves show results over training time, averaged across seeds.
    }
    \label{fig:gender_classification}
\end{figure}

\paragraph{Method}
Previous work has shown that gender (and resulting biases) are encoded by only a small number of units \cite{DBLP:conf/nips/VigGBQNSS20, decao2021sparse}. %
We build on this work and examine what stages a model undergoes in order to obtain such a local representation of gender. %
For all the saved checkpoints of our LSTM models we train a \textbf{linear classifier} based on the word embeddings of 82 gendered word pairs (e.g. \textit{he}-\textit{she}, \textit{son}-\textit{daughter}, the full list is shown in Appendix \ref{app:wordlists}.). 
The classifier is trained with L2 regularisation on an 80/20 train/test split. 
We utilise the distance to the resulting decision boundary as a proxy for the gender subspace of the model.
With the resulting set of classifiers, we conducted several experiments to gain insights into \textit{how} gender is represented.

\subsection{How localised is the representation of gender?}
\label{sec:locality}
The results of the gender classification task are shown in Figure~\ref{fig:gender_classification}.
The performance on the test corpus (\textcolor{orange}{orange curve}) can be seen to be increasing gradually over time, already reaching around 85\% at the end of the first epoch, and settling at around 93\% after 3 epochs of training.
Furthermore, and in line with results from other studies, we find that the representation of gender is very localised:
a single unit in the embeddings dominates the representation of gender, which we call the \textbf{gender unit}.

To quantify how well this unit captures gender, and how this quantity changes over time, we  train a new classifier that uses \textit{solely} the gender unit in the embeddings. 
Results for this experiment are shown as the \textcolor{green_plt}{green curve} in Figure~\ref{fig:gender_classification}. We find that in the initial stages, this classifier performs at chance level. 
After this stage, a surprisingly gradual increase in accuracy takes place, and after around 4 epochs of training the model settles at a local gender representation with an accuracy of around 90\%.

The single gender unit is thus able to capture gender almost as well as the classifier that had access to the full embedding. 
To investigate to what extent this unit is special in capturing gender compared to the other embedding axes we also train classifiers in which the gender unit has been \textit{removed} (\textcolor{red}{red curve}). 
It can be seen that in the initial stages this classifier performs on par with the full classifier. 
However, along the course of epoch 1 it slowly starts to deteriorate; in epoch 2, it is even being surpassed by the single gender unit classifier.  %
These three curves show that the model has concentrated the majority of gender information into a single unit, but that part of it is still distributed over the remaining axes.

To see how the gender unit develops over time we compute whether or not the dominant gender unit is the same at different time points (see Figure~\ref{fig:axis_overlap} in the appendix). 
After \textasciitilde 30\% of epoch 1 the model has settled on what the main unit is going to be to represent gender on.
Prior to that point the model undergoes a phase in which it alternates between several gender units, none of which are equal to the final gender unit. 
Even though the model has already settled on the final gender unit at an early point, it still takes more than a full epoch of training before it has arranged its word embedding space in such a way that gender is captured optimally by that unit.

We utilise these findings to define three distinct phases that a model undergoes to form its  representation of gender: i) the \textbf{formation phase}, in which the model is exploring a suitable gender representation; ii) the \textbf{consolidation phase}, after around 30\% of epoch 1, in which the model gradually restructures its space around the newly found gender representation; iii) the \textbf{specialisation phase}, after around 3 epochs, in which the model amplifies the gender signals that have been formed in the previous phase.

\begin{figure*}[bht]
    \centering
    \begin{subfigure}[b]{0.329\textwidth}
    \includegraphics[width=\textwidth,trim=0cm 0cm 22.2cm 7.5cm, clip]{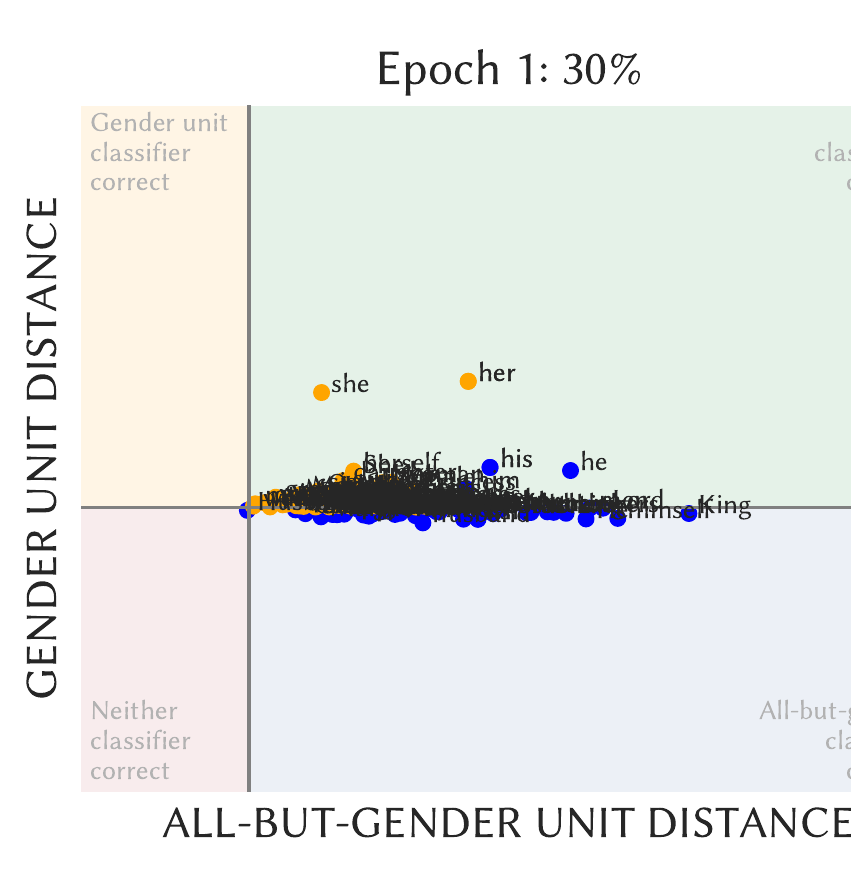}
    \end{subfigure}
    \hfill
    \begin{subfigure}[b]{0.329\textwidth}
    \includegraphics[width=\textwidth,trim=11.1cm 0cm 11.1cm 7.5cm, clip]{figures/threshold_tokens.pdf}
    \end{subfigure}
    \hfill
    \begin{subfigure}[b]{0.329\textwidth}
    \includegraphics[width=\textwidth,trim=22.2cm 0cm 0cm 7.5cm, clip]{figures/threshold_tokens.pdf}
    \end{subfigure}
    \caption{Gender information encoded on the dominant gender unit (plotted vertically) mainly serves to distinguish female words from other words; gender information encoded in all other units (plotted horizontally) mainly serves to distinguish male words. Shown are distances from the decision boundary for the gender-unit-only classifier and for the gender-unit-removed classifier for each word, at three different time steps during training.
    }
    \label{fig:scatter-plots}
\end{figure*}

\subsection{Which words drive the organisation of the gender representations?} %
Next, we examine which tokens play a vital role in the shaping of a model's gender representation. 
Soon after the start of training, %
certain embeddings start to reflect %
(linguistic) features such as gender. 
Slowly, the model %
forms a more general notion of gender, aligning other (gendered) tokens with the initial set of gendered tokens that drove the learning process.
We utilise the decision boundary distance to examine which tokens play an early role in the development of gender.
We do this for two types of classifiers: (i) the single gender unit classifier that has been explored in the previous experiments, and (ii) the classifier that utilises all but the gender unit.

The result for this procedure is shown in Figure~%
\ref{fig:scatter-plots}.
We see a striking pattern emerging here: the development of the dominant gender unit is strongly driven by female tokens, whereas male tokens dominate the development of gender information that is distributed across all other dimensions. 
This is in line with earlier work that showed that masculinity acts as the default gender class for a language model \cite{DBLP:conf/conll/JumeletZH19}.
A model will only prefer the prediction of a feminine token once it has encountered explicit evidence for it, and it is able to do so by channelling this information through a localised dimension.

%% file: sections/experiment2.tex
\section{The evolution of gender bias}
\label{sec:experiments-characterizing}

Building on the last section, we now turn our attention to gender bias, i.e. the association of gender with words that are not explicitly gendered. %
Specifically, inspired by previous work \cite{caliskanSemanticsDerivedAutomatically2017,rudinger2018GenderBiasCoreference,zhao2018GenderBiasCoreference,webster2020MeasuringReducingGendered} we consider the gender bias for 54 occupation terms (see Table \ref{tab:wordlists}(c) in the appendix).

\subsection{From gender representation to gender bias}
\label{sec:gender_bias}

We follow \citet{ravfogel2020NullItOut} and use a \emph{support vector machine} to find the optimal linear decision boundary between 18 unambiguously feminine and masculine words (also used by previous work \cite[e.g.][]{bolukbasi2016ManComputerProgrammer,ethayarajhUnderstandingUndesirableWord2019,ravfogel2020NullItOut}
\footnote{We leave out the word pair (`guy', `gal'), as we have noticed better results without the word pair. \citet{ethayarajhUnderstandingUndesirableWord2019} and \citet{du2021assessing} warn that including low-frequency words can negatively impact the bias measure, which we suspect is the case here.}), of which the orthogonal axis serves as the primary gender subspace, $\vec{g}$.
Given this subspace $\vec{g}$, gender bias (w.r.t. the gender-neutral norm) can be defined using the scalar projection of every input embedding, $\vec{w}$, onto the subspace, see Equation \ref{eq:IE_bias}.\footnote{
Please note that the gender subspace we define here is closely related to the approach in Section \ref{sec:experiments-gender-representation} for identifying the \emph{gender unit}.
Even though the classifier for finding the \emph{gender subspace} is only trained on a subset of the gendered word-list used in the previous section---which is done to match previous work more closely \cite[e.g.][]{bolukbasi2016ManComputerProgrammer,ethayarajhUnderstandingUndesirableWord2019,ravfogel2020NullItOut}---we find that the decision boundaries of both approaches correlate very strongly and that the observations on the locality of gender information are relevant here as well.}

\begin{equation}
    \label{eq:IE_bias}
    \text{bias}_\text{IE}(w) = \langle \vec{g}, \vec{w} \rangle
\end{equation}

The resulting scalar value quantifies the strength of the bias, while the sign indicates the direction on the female-to-male axis.
In the rest of the paper, we refer to this bias as the input embedding (IE) bias.

When studying the average input embedding bias for the non-gendered occupation terms, we observe a steady (absolute) increase over the course of training, with the strongest growth in the first half of epoch 1, and a levelling off in the last 20 epochs (we refer to Figure~\ref{fig:avg_bias} in the appendix).

Does this spreading out of occupation terms along the gender dimension correlate with a bias in downstream behaviour? 
For the purpose of this paper, we use the \emph{semantic textual similarity} task adapted for gender bias \cite[STS-B,][]{webster2020MeasuringReducingGendered}, which, as is common in the literature, measures bias on a carefully created collection of sentences (a `challenge set').
This task contains 276 template sentences $t\in T$, where for each occupation $o$ that sentence either starts with that occupation, "man", or "woman", resulting in a triplet ($t(o)$, $t(\text{``man''})$, $t(\text{``woman''})$).
One of the sentence triplets is, for example, \textit{``A man/woman/janitor is playing the guitar''}.
The gender bias for occupation $o$ is calculated as the average difference in similarity with the sentence starting with "man" compared to the sentence starting with "woman", see Equation~\ref{equ:similarity}. We use the cosine similarity of the last hidden states of our LSTM model as a proxy for the semantic similarity, to avoid training an additional semantic similarity classifier and making the relationship to the earlier stages of the language modelling pipeline less interpretable.

\begin{equation}\label{equ:similarity}
\begin{split}
    \text{bias}_\text{STS-B}(o) =
    \frac{1}{|T|} \sum_{t \in T} \text{similarity}(t(o), t(\text{``man''})) \\
    - \text{similarity}(t(o), t(\text{``woman''})
\end{split}
\end{equation}

With this measure for downstream bias in our hands, we can return to the questions whether there is a relation between the dynamic behaviour of bias we observed in the input embeddings and the bias in downstream behaviour. 
The answer is a qualified yes. 
We find that the progression of bias in the STS-B task grows very rapidly in the first few training batches, is extremely variable during epoch 1, and does grow to a level of around 0.3 in the second half of epoch 1 (see Figure~\ref{fig:avg_bias} in the appendix). 
It then remains around that point for the remaining 39 epochs.

Moreover, while the \emph{change} in the metrics is clearly no longer correlated from halfway epoch 1, at each time slice we do find a fairly strong correlation between the two measures across the vocabulary of interest. E.g., at epoch 40 we find a correlation of 60\%, indicating that the input embedding bias scores for \emph{nurse}, \emph{receptionist}, \emph{engineer}, \emph{architect}, \emph{mechanic}, etc. are fairly predictive of the downstream bias scores for STS-B sentences containing these words.

\begin{figure*}
    \centering
    \includegraphics[width=\textwidth]{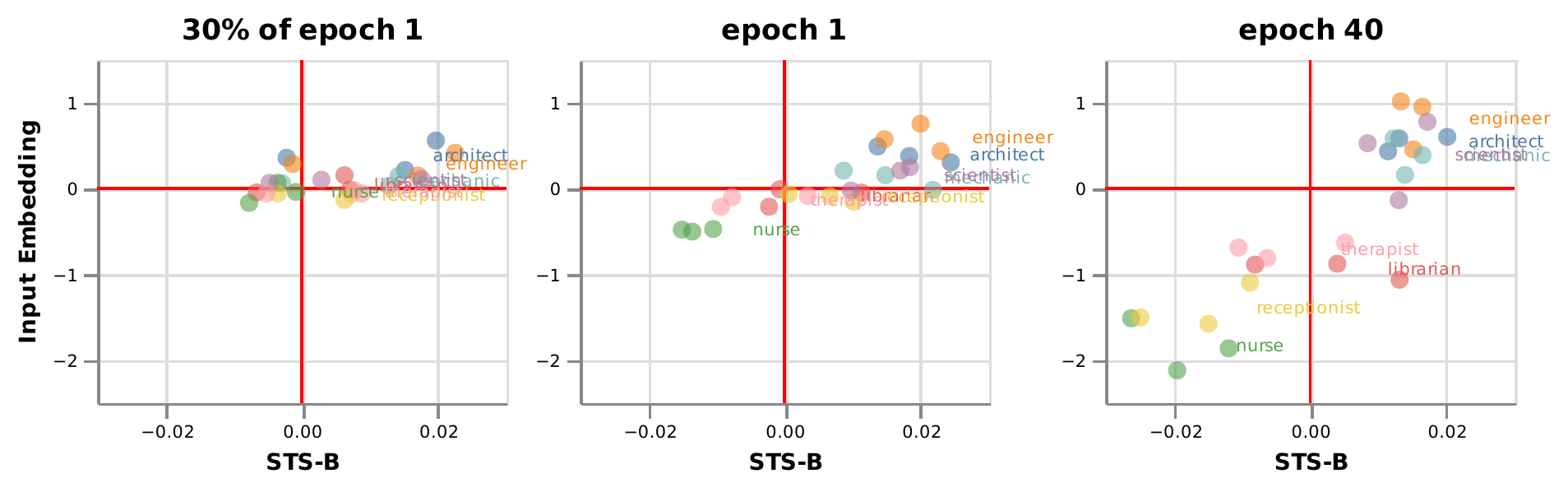}
    \caption{For three different points in time, we show the input embedding bias and STS-B bias for a selected few occupation words, averaged over the three different random seeds. The occupation terms visualised in this figure are ``receptionist'', ``nurse'', ``librarian'', ``therapist'', ``mechanic'', ``engineer'', ``scientist'', and ``architect'', which we have found to display strong biases for both bias metrics.} 
    \label{fig:bias-progression}
\end{figure*}

A similar observation can be made when looking at individual occupation words (Figure~\ref{fig:bias-progression}). 
Here as well do we find that the input embedding and STS-B bias are correlated. 
For instance, both bias measures broadly capture a strong male bias for ``engineer'', while ``nurse'' and ``receptionist'' have a strong female association for both representations.
Complementary to this, we find that for both the input embeddings and STS-B, some words show these biases much sooner than other words.
For the word ``nurse'', for example, a female bias can be found earlier during training than for ``receptionist', even though both have a strong female bias after training.
We hypothesize that this reflects the differences in their dataset statistics.
For instance, we find that ``nurse'' occurs 783 times, while ``receptionist'' only 66 times.
On top of that, ``nurse'' also has a higher PMI association with female gendered words (we explain the PMI association in more detail in Section~\ref{sec:dataset_statistics}).

However, the fact that the correlation between the two metrics is not higher than $0.6$ highlights that there still are some important differences.
First, we notice that the STS-B bias is noisier than its input embedding counterpart in the first epoch, which is not a surprise given that the language modelling relies on contextual information and is measured on a relatively small set of examples. %
More importantly, however, we observe in Figure \ref{fig:bias-progression} that the gender bias is heavily skewed towards a female bias for the input embeddings, but this asymmetric pattern is not as apparent for the STS-B task.
It seems that this asymmetry gets masked at the level of the downstream task, but the underlying cause is still asymmetric, which is relevant when considering countermeasures.
We will come back to the asymmetry in gender bias in Section~\ref{sec:intervention_experiment}.

\subsection{Relating gender bias back to dataset statistics}
\label{sec:dataset_statistics}

So far, we have seen that the way gender is represented in gendered words helps us understand how gender bias is represented in the input embeddings of non-gendered words, and how these representations change over time. Moreover, we have seen that the used bias metric at the level of these input embeddings is fairly predictive of the downstream bias measured through STS-B. We now turn our attention to the question of how and why non-gendered words get mapped to the emergent gender axes of the language model. For this, we examine how well the model biases correlate with dataset features and external U.S. labour statistics with the ratio of male and female workers for each occupation (see Appendix \ref{app:labour_stats}). We will not be able to give a firm answer to this question, as neural models are capable of learning from more sophisticated, and perhaps implicit, features of the dataset than we consider, but there are still some interesting patterns we can observe.

Following others \cite{zhao2019GenderBiasContextualizeda, tan2019AssessingSocialIntersectional,fast2016shirtless,gao2020pile}, we examine the word-count statistics for the dataset in our experiments. %
We consider two statistics, namely (i) the word counts and (ii) the pointwise mutual information (PMI) with a set of 18 gendered words (see Table \ref{tab:wordlists}(a,b) in Appendix \ref{app:wordlists}).
The PMI statistic is defined as given in Equation~\ref{equ:pmi}, where $p(x)$ indicates the probability of word $x$, which we estimate by the word count $c(x)$. The joint probability $p(x,y)$ is estimated with the co-occurrence count for words $x$ and $y$, for which we use a sliding window of 35 tokens that is equal to the BPTT window of our LSTM models.
In our case, $x$ is an occupation and $y$ the set of gendered words (either female or male words, indicated by subscript $\venus$ and $\mars$, respectively).
We also combine the PMI statistics for the two genders to capture an aggregate association, where
$PMI_{\mars - \venus} = PMI_{\mars} - PMI_{\venus}$.

\begin{equation}
\label{equ:pmi}
    \text{PMI}(x,y) = \log \frac{p(x,y)}{p(x)p(y)} = \log \frac{c(x,y)}{c(x)c(y)}
\end{equation}

We first check how these statistics are correlated with each other, looking only at the dataset (independently of the language model). We find that $PMI_{\venus}$ ($-0.23$) correlates fairly well with the labour statistics, and more strongly than $PMI_{\mars}$ ($0.12$). In other words, female gendered words (``she'', ``her'', ``woman'') in the vicinity of an occupation term, are weakly predictive of the percentage of female or male workers in that occupation, while male gendered words reveal less\footnote{This is in line with the often observed \emph{male-as-norm} phenomenon in language: the male category is used more generally, while female gendered words are more specific for indicating that particular gender \cite{danesi2014dictionary}.}. The highest correlation, however, is obtained with an aggregate of the two PMI measures, $PMI_{\mars-\venus}$ ($0.33$).

\begin{figure*}
    \centering
    \includegraphics[width=\textwidth]{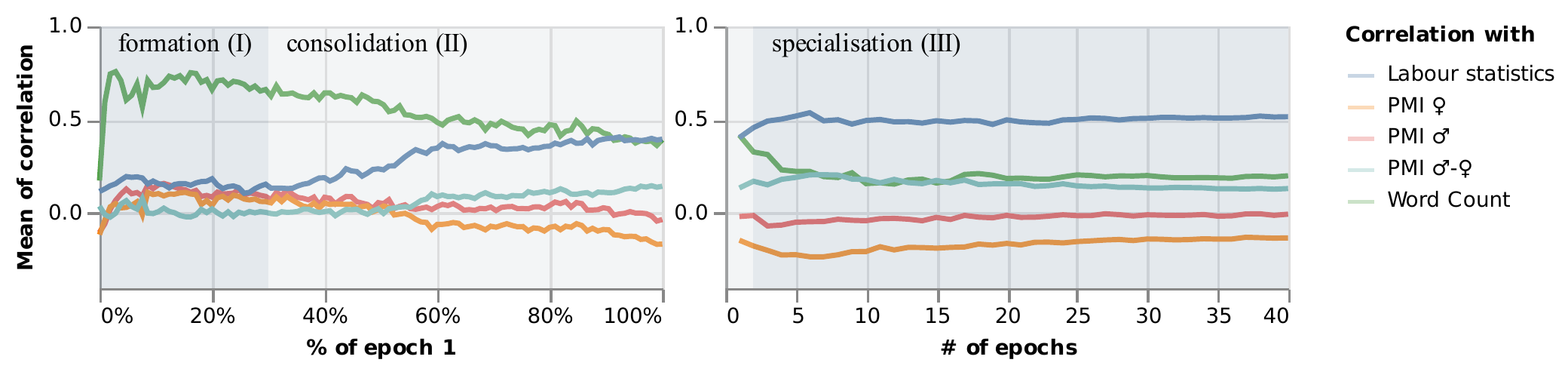}
    \caption{Pearson correlation of the input embedding bias scores for the occupation terms with the different dataset features (word count and PMI) and labour statistics, and how these change during training time. %
    }
    \label{fig:corr_bias}
\end{figure*}

Partitioning the training period in the three phases for gender representations that we identified in the previous section, we see an interesting pattern of results for the correlation with the input embedding bias (see Figure~\ref{fig:corr_bias}).  %
In the \emph{formation} phase, all correlations are low, except for the correlation with word count; i.e., high bias scores are best predicted by simple frequency of the term.

In the \emph{consolidation} phase, word count starts losing its predictive power for bias, and the correlation with the labour statistics starts building up, reaching approximately 40\% by the end of the \emph{consolidation} phase and remaining there throughout the \emph{specialization} phase. Note that the labour statistic is external; the language model only has access to statistical patterns that are reflected in the input text. We do not know which text statistic mediate the formation of this correlation, but it is interesting that the steepest growth of the labour statistic correlation, coincides with the aggregate PMI-measure $PMI_{\mars-\venus}$, taking dominance over both female and male specific measures.

\subsection{Summary}
The projection on the gender subspace from Section~\ref{sec:experiments-gender-representation} finds plausible gender associations with different occupations, and %
we observe that the input embedding bias measure is predictive of the bias that we measured in the STS-B task. However, the correlation is far from perfect, and there are some interesting differences between both measures with respect to gender asymmetry.
We also saw that the input embedding bias can be to some extent related to statistics in the dataset, although the language model clearly picks up on many more sources of information on gender association than can be captured by measures like PMI.

Finally, each of the correlations we studied shows interesting dynamics over training time. 
We see that our measures for input embedding bias and downstream bias grow together during the formation phase, but decouple during the consolidation phase; %
that word count is dominant in the formation phase, but becomes a progressively less important data statistic in later phases; 
and that the aggregate PMI measure gives better correlations than separate $PMI_{\mars}$ and $PMI_{\venus}$ about halfway the first epoch.

%% file: sections/experiment3.tex
\section{Diagnostic intervention: changing downstream bias by changing embeddings}
\label{sec:intervention_experiment}

So-far, our analyses have all been correlational. 
In this section, we aim at establishing a causal role for the representations of gender and gender bias that we have described in the previous two sections. 
We do so by \emph{intervening} on the input embeddings, using the debiasing method \textbf{Iterative Null-space Projection} of \citet{ravfogel2020NullItOut}.
In each debiasing step, a gender subspace is identified (as discussed in the previous section), after which all word vectors are projected on its null-space to remove this gender information.
The authors show that performing a null-space projection once is not sufficient for removing bias completely.
However, repeating this procedure multiple times turns out to be an effective mitigation strategy, without an overall decay of the embeddings \cite{ravfogel2020NullItOut}. %
In our experiments, we denote the number of null-space projections as $k$.

We apply this method to the input embeddings, and measure the effects on the downstream behaviour, again using the STS-B task. 
Our goal is not, in the first place, practical (i.e. to end up with an unbiased language model), but rather diagnostic: shedding light on the nature of the representation of gender and gender bias, and the way they influence the behaviour of the model. \
Ultimately, we hope that our analysis allows us to draw conclusions about the conditions under which debiasing input embeddings (using this particular method) might be an attractive strategy to mitigate bias in contextual word embeddings.

\subsection{Comparing the effect of debiasing across training time}

\begin{figure}[htb]
    \begin{subfigure}[b]{0.49\columnwidth}
    \includegraphics[width=\textwidth,trim=0cm 0cm 12.5cm 0.65cm, clip]{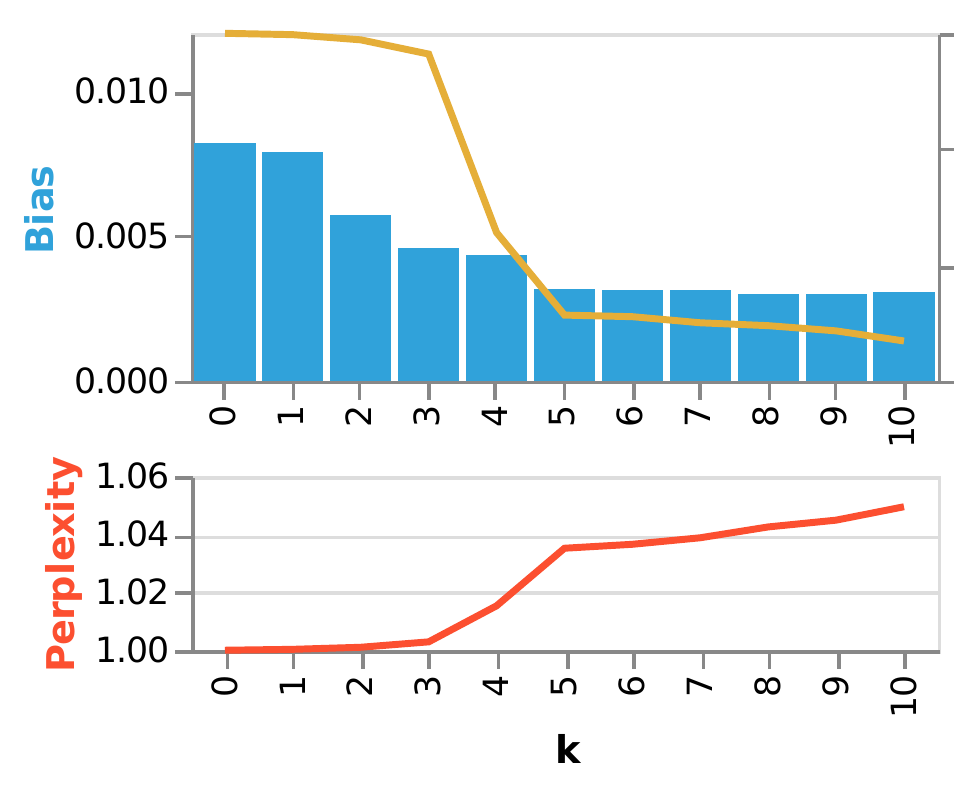}
    \end{subfigure}
    \hfill
    \begin{subfigure}[b]{0.49\columnwidth}
    \includegraphics[width=\textwidth,trim=12.65cm 0cm 0cm 0.65cm, clip]{figures/snapshot_debias.pdf}
    \end{subfigure}%
    \caption{The average STS-B bias, RSA similarity with the original embeddings ($k=0$), and perplexity values after $k$ debiasing steps for two points in time: epoch 1 (\textit{left}) and 40 (\textit{right}).
    The perplexity is normalised with respect to the original language model before debiasing.
    Please note that the starting perplexities are different for epoch 1 and 40.}
    \label{fig:bias_perplexity_snapshots}
\end{figure}

We perform the Iterative Null-space Projection 10 times on the input embeddings of our language models, at two different points in training: after the first and last epoch.
The representation of gender is likely to comprise multiple linear components \cite{ravfogel2020NullItOut}, with the most dominant one being the \emph{gender unit} from Section \ref{sec:experiments-gender-representation}.
By repeating the debiasing procedure multiple times, we can learn more about this underlying representation as well as explore how these change during training time.
The results for this experiment are shown in Figure \ref{fig:bias_perplexity_snapshots}.
We measure the downstream bias using STS-B for the original embeddings, as well as for each of the ten iterations of the debiasing algorithm. 
Moreover, we also measure the quality of the language modelling using the standard perplexity metric.
Finally, we measure qualitative changes in the topological organisation of the semantic space of the occupation and gendered words\footnote{See the word-lists in Appendix \ref{app:wordlists}.}, by measuring Representational Similarity \cite{kriegeskorte2008representational} between the original model and the debiased models.

Figure \ref{fig:bias_perplexity_snapshots} shows a number of important effects. First, we see that there is a visible decrease of the measured bias after debiasing the input embeddings. And, importantly, both perplexity and Representational Similarity show only minor changes up to three debiasing iterations. Performance of the language model starts to diminish at four debiasing steps, and decrease further at five steps and more. These results are in line with our earlier findings about how gender is represented: mostly along the dominant gender unit (Section \ref{sec:experiments-gender-representation}), but with gender information also encoded in the rest of the embedding space, and mostly encoded in such a way that it can be decoded using linear classifiers.

Strikingly, debiasing is much more effective at epoch 40 (the end of training, and the end of the \emph{specialisation} phase), than at epoch 1 (the end of the \emph{consolidation} phase). At epoch 40, the average bias of the model is worse before debiasing, but much better after debiasing, reaching a bias score of just above $0.01$. These results agree with our earlier observations that the gender information is encoded more locally during training, which would be easier to remove effectively and selectively. 
For our specific setup, three debiasing iterations seems to be a sweet-spot, where the perplexity increase is still minimal and the debiasing effect is strong.

\subsection{Asymmetry in debiasing female and male bias}
\begin{figure*}
    \centering
    \begin{subfigure}[b]{0.49\textwidth}
    \centering
    \includegraphics[width=\textwidth]{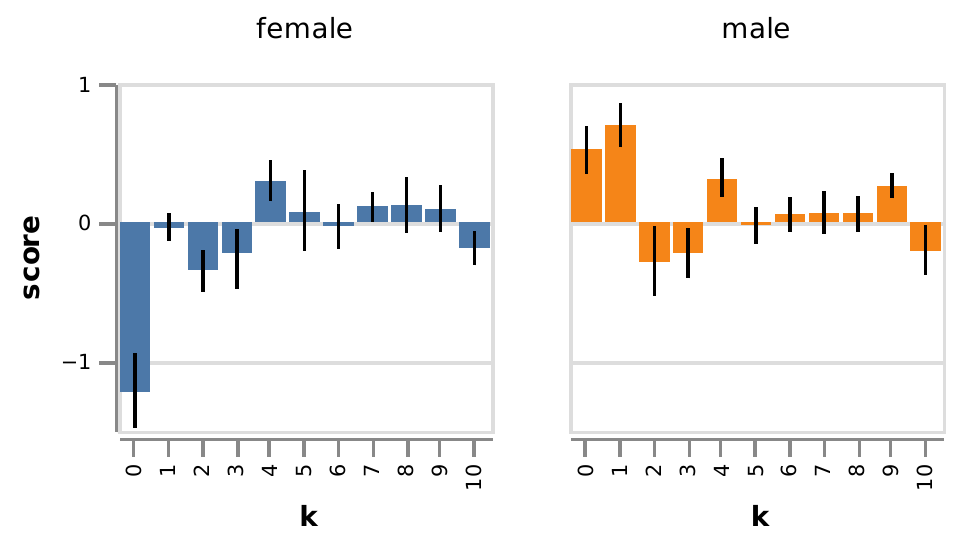}
    \caption{Input embedding bias}
    \end{subfigure}
    \hfill
    \begin{subfigure}[b]{0.49\textwidth}
    \centering
    \includegraphics[width=\textwidth]{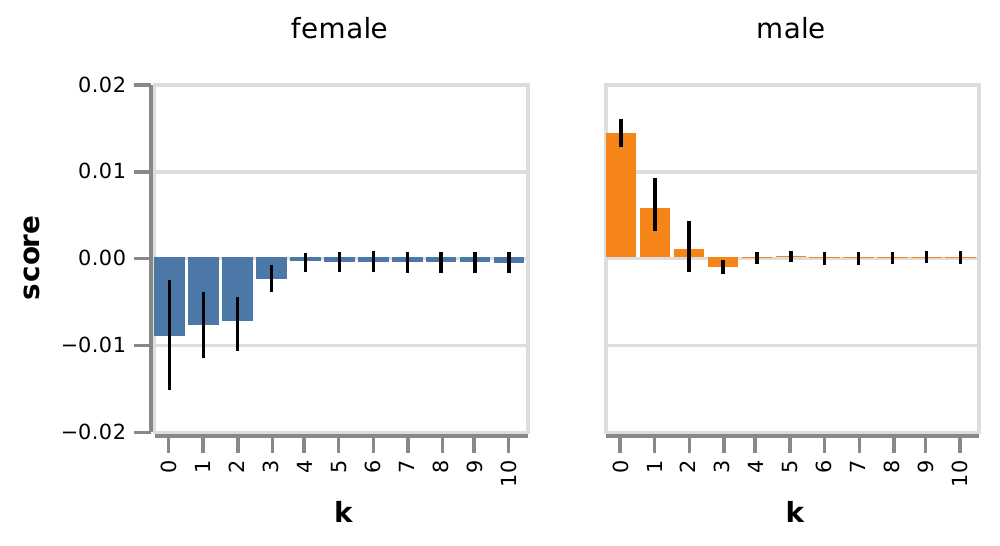}
    \caption{STS-B bias}
    \end{subfigure}
    \caption{Effect of debiasing the language model at epoch 40 on the bias scores for a list of male- and female biased occupations. Based on the bias scores for IE and STS-B bias, we chose ``receptionist'', ``nurse'', ``librarian'', and ``therapist'' for the female words and ``mechanic'', ``engineer'', ``scientist'', and ``architect'' for the male word-list.}
    \label{fig:effect_debiasing}
\end{figure*}

To get a more fine-grained picture of how debiasing the input embeddings affects downstream bias, we also consider the effect on the female and male bias separately, as we expect some asymmetry from our earlier observations in Sections \ref{sec:experiments-gender-representation} and \ref{sec:experiments-characterizing}.
Figure \ref{fig:effect_debiasing} displays the bias scores for a set of female and male biased occupation words for the fully trained language model after $k$ debiasing steps. For this figure, we measure bias both on the input embeddings and in the STS-B task.
We can see that a single debiasing step already has a visible effect on both 
bias measures.

Interestingly, when we consider the input embeddings, we see a strong reduction of the female bias. In contrast, we even observe an increase of the average male bias after one debiasing step.
Only after another few steps do we see that both the male and female bias get reduced more significantly. 
These results are related to our earlier observations in Section \ref{sec:experiments-gender-representation} about gender asymmetry: the dominant gender unit is used primarily to encode the feminine feature of words, while masculine word information is more distributed over the rest of the input embedding space. %
We also observe a slight increase in bias after $k>6$, which we attribute to the bias metric being sensitive to noise in the absence of an actual linear gender representation.\footnote{In actual applications of Iterative Null-space Projection this is less of a problem, since you typically stop debiasing if the accuracy of classifier is close to random.}

The STS-B bias, however, shows a different behaviour.
Debiasing the input embeddings clearly has an effect on the downstream behaviour, but debiasing once has a larger effect on the \emph{male bias} instead.
It takes more than two debiasing steps before both the female and male bias is reduced.
Interestingly, we found earlier that three debiasing iterations is a sweet spot, but we have no satisfying explanation for why especially the male bias is reduced in the first iteration.

\subsection{Summary}
We conclude that there is a causal effect of the gender representation in the input embeddings on the downstream bias.
First, we find that the Iterative Null-space Projection is surprisingly effective and that three debiasing steps result in a bias reduction with minimal harm to the perplexity of the language model and topological representation of the embedding space.
This reflects our earlier finding that gender information is encoded very locally, but also suggests that the model relies a lot on this \emph{linearly decodable} gender representation.
Secondly, we find that observing the effect on male and female biased occupation terms separately shows an asymmetry for both the input embedding and STS-B bias.
While the asymmetry towards female bias in the input embeddings can be explained by our earlier observations in Section \ref{sec:experiments-gender-representation}, we are not sure why removing this information affects especially the male bias in the contextual embeddings.
More work is needed to explore possible explanations for this incongruity as it can have important consequences for certain types of mitigation strategies.

%% file: sections/discussion.tex
\section{Discussion}
\label{sec:discussion}

Although in our experiments we have restricted ourselves to gender bias in English, we believe
our results have relevance for the broader study of bias in language models. 
More concretely, this paper contributes to the ongoing research on bias in language models in three ways: 
we shed light on the question of how the internal representation of the model relates to its downstream bias, we show that studying the dynamic nature of bias can be illuminating, and we point out that there are potential asymmetries in the underlying bias representation that researchers should be aware of. 

\subsection{Relationship internal representations and downstream bias}

When deciding on a bias mitigation strategy, it is crucial to understand the relation between the internal representations of the language model and the bias in downstream tasks. This is because successful debiasing of the internal representations will likely generalise over many downstream tasks, but this strategy is only viable if the internal representation that is manipulated is causally linked to the downstream behaviour of the model. Whether this causal connection exists, however, might differ from case to case.

In a study looking at static word embeddings (not language models, as we do here) and a number of different downstream tasks, \citet{goldfarb2020intrinsic} find no correlation between the bias in the embeddings and in the downstream tasks. In constrast, 
\citet{ravfogel2020NullItOut} find that debiasing embeddings can be effective in reducing racial bias in a sentiment classifier.
More closely related to our setup of investigating gender bias in a language model, \citet{vig_causal_2020} and \citet{decao2021sparse} actually show that gender information can be stored in or mediated by a small part of a language model by selectively changing neuron activations and analysing the effect on the output.
\citeauthor{decao2021sparse} even study the parameters of the same LSTM architecture that we consider in this paper. 

In line with these findings, we also observe that gender information is represented very locally in the input embeddings of the LSTM language model. Furthermore, we find that manipulating bias in the input embeddings does indeed affect downstream bias, adding evidence for a causal relation between this particular level of internal representation and downstream behaviour of the model. %
However, we should add here that whether one finds such a connection might strongly depend on the choice of bias metric, model architecture, and downstream task, and furthermore depends on the particular learning phase a language model is in with respect to a particular type of bias.

\subsection{Different phases in the evolution of gender}
Based on our finding we can distinguish three phases in the evolution of gender representation and gender bias: (i) formation, (ii) consolidation, and (iii) specialisation. We saw that our measures for bias and the method for bias mitigation behave differently in these different phases, which appears to be connected to how locally gender information is represented in the internal representations of the model. Only if the relevant information is concentrated in a particular part of the model and linearly decodable, can we reliably and selectively remove gender information without hurting the overall language model performance.

This observation might not be so important when thinking about gender bias in current large language models, as the sheer scale of the datasets that these models are trained on and the high frequency of gendered words makes it very likely that they have progressed far into the 'specialisation phase' with respect to their representation of gender. However, it could matter when considering other types of biases, where the words and phrases driving the birth of these biases may be much less frequent.
Hence, even in large language models trained on several orders of magnitude more data than the language model we used in this study, the relevant representations for other biases might very well still be in something equivalent to our `formation' or `consolidation phase'.
Indeed, work on studying the effect of fine-tuning has shown that the manifestation of bias can still change significantly in pre-trained models \cite{choenni2021stepmothers,webster2020MeasuringReducingGendered}. %

\subsection{Asymmetry in the gender representation}
Gender asymmetries are regularly observed in word frequencies and co-occurrences in datasets \cite[e.g.][]{zhao2019GenderBiasContextualizeda,tan2019AssessingSocialIntersectional,wagner2016women} and in language use in general \cite[e.g. the ``male-as-norm bias'',][]{danesi2014dictionary}.
Interestingly, we also observed a strong asymmetry in how gender bias is represented in the input embeddings, but we did not see the same asymmetry in the downstream task.
This could have consequences for how mitigation strategies should be evaluated.
When debiasing the model while being unaware of the underlying representation, one could disproportionately harm one group more than another. This could lead to the introduction of a new form of bias.
In developing and evaluating mitigation strategies, it is therefore important to do a thorough analysis of the representation of bias present in the NLP system and how certain social groups could be affected disproportionally if not accounted for.

%% file: sections/conclusion.tex
\section{Conclusion}
While there is a lot of important work on detecting and mitigating undesirable biases in language models, we still lack a good understanding of the mechanisms underlying the biased behaviour. The goal of this study was to take a step back and analyse the birth of bias in language models.
To this end, we present a temporal investigation of how an English LSTM language model learns a representation of gender in the input embeddings and how this affects downstream biased behaviour.

There are many interesting directions for future research. An important open question is, for instance, how intrinsic representations of bias relate to other downstream tasks that may be closer to real-world systems where the representational and allocative harms to social groups are more clear \cite{blodgett2020language}.
For future work, we also plan to do further investigations on how our training dynamics analysis may generalise to other undesirable social biases, model architectures, training corpora, and downstream tasks, as well as other possible representations in the internal states of the language model that are useful for understanding bias.
Furthermore, the robustness of our results with respect to different random initialisations of the language model should be checked \cite{webster2020MeasuringReducingGendered,damour2020UnderspecificationPresentsChallenges}.

In this paper, we take a step towards a more thorough understanding of the evolution of bias in language models across the different stages of the language modelling pipeline. Hopefully, it will inspire more work on the dynamic behaviour of language models, with respect to bias, but also other still poorly understood features of these models.

%% file: sections/appendix.tex
\appendix
\label{sec:appendix}

\section{Labour statistics}
\label{app:labour_stats}

In this work, we use the US Bureau of Labor statistics on the percentage of female workers \cite{caliskanSemanticsDerivedAutomatically2017} for comparison with the gender bias in the language modelling pipeline (see Table \ref{tab:labour_stats}).\footnote{\url{https://github.com/rudinger/winogender-schemas/blob/master/data/occupations-stats.tsv}}
Please note that the ordering of this list can be reversed when computing correlations.

\begin{table*}[h]
    \centering
    \begin{tabular}[t]{lr}
    \toprule
       occupation &  \% female \\
    \midrule
      pathologist &           97.50 \\
        secretary &           94.60 \\
      hairdresser &           94.20 \\
     receptionist &           90.60 \\
            nurse &           89.58 \\
        librarian &           83.00 \\
          planner &           77.60 \\
        therapist &           76.70 \\
     practitioner &           74.79 \\
          cashier &           72.50 \\
          teacher &           71.00 \\
         educator &           70.80 \\
     psychologist &           70.30 \\
            clerk &           69.53 \\
        counselor &           66.48 \\
         examiner &           62.46 \\
       instructor &           62.30 \\
            baker &           60.80 \\
     veterinarian &           60.50 \\
        bartender &           59.80 \\
          auditor &           59.70 \\
       accountant &           59.70 \\
       pharmacist &           57.00 \\
       dispatcher &           56.30 \\
           broker &           55.50 \\
    administrator &           54.86 \\
     investigator &           45.15 \\
    \bottomrule
    \end{tabular}
    \begin{tabular}[t]{lr}
    \toprule
    occupation &  \% female\\
    \midrule
    scientist &           41.94 \\
   specialist &           41.35 \\
   technician &           40.34 \\
   supervisor &           38.64 \\
      manager &           38.51 \\
       worker &           37.92 \\
       doctor &           37.90 \\
      advisor &           37.90 \\
    physician &           37.90 \\
      surgeon &           37.90 \\
      chemist &           36.10 \\
       lawyer &           34.50 \\
      janitor &           34.30 \\
    paramedic &           32.90 \\
      officer &           30.42 \\
    architect &           20.81 \\
         chef &           19.60 \\
   programmer &           18.35 \\
     engineer &           10.72 \\
    machinist &            6.70 \\
    inspector &            6.40 \\
      painter &            5.70 \\
  firefighter &            3.50 \\
  electrician &            2.30 \\
    carpenter &            2.07 \\
     mechanic &            1.80 \\
      plumber &            0.70 \\
    \bottomrule
    \end{tabular}
    \caption{US Labour Statistics with the percentage of female workers for the occupations we consider in gender bias analysis of the LSTM language model. Any differences with previous work are due to some occupations being left out, as these do not occur in the model vocabulary.}
    \label{tab:labour_stats}
\end{table*}

\section{Dataset statistics}
In Section \ref{sec:experiments-characterizing}, we score the occupation words with various dataset features and rank these with the labour statistics. The results can be found in Table \ref{tab:wordcount_statistics}.

\begin{table*}[h]
\centering
\begin{tabular}{lr}
\toprule
Dataset feature         &    Pearson correlation         \\
\midrule
Word Count              &    0.108785 \\
PMI$_{\mars}$             &    0.116814 \\
PMI$_{\venus}$            &   -0.234644 \\
PMI$_{\mars - \venus}$    &    0.333687 \\
\bottomrule
\end{tabular}
\caption{Pearson correlation with labour statistics for occupation words.}
\label{tab:wordcount_statistics}
\end{table*}

\section{Wordlists}
\label{app:wordlists}
We use two sets of word-lists in the experiments of Sections \ref{sec:experiments-gender-representation}, \ref{sec:experiments-characterizing}, and \ref{sec:intervention_experiment}.
The first word-list used in Section \ref{sec:experiments-gender-representation}, contains a list of 82 gendered word-pairs (also considering capitalised and pluralised versions in the model vocabulary), as shown in Table~\ref{tab:gendered_wordlists}.
Then, in Sections \ref{sec:experiments-characterizing} and \ref{sec:intervention_experiment}, we use a subset of the previous gendered word-pairs that is more similar to what is used in previous work for finding a \emph{gender subspace} \cite[e.g.][]{bolukbasi2016ManComputerProgrammer,ethayarajhUnderstandingUndesirableWord2019,ravfogel2020NullItOut} and can be found in Table \ref{app:wordlists}.
This last table also contains a list of 54 occupation words for studying gender bias, which corresponds to the list in Table \ref{tab:labour_stats}.
We have indicated the overlap between the two word-lists in bold for reference.

\begin{table*}[h]
\begin{tabular}{p{2.5cm}|p{12cm}}
\toprule
\textbf{Type} & \textbf{Words}                                                                                  \\ \midrule
Male          & man, boy, he, father, son, male, his, himself, John\\
Female        & woman, girl, she, mother, daughter, female, her, herself, Mary\\ 
Occupations   & technician, accountant, supervisor, engineer, worker, educator, clerk, counselor, inspector, mechanic, manager, therapist, administrator, receptionist, librarian, advisor, pharmacist, janitor, psychologist, physician, carpenter, nurse, investigator, bartender, specialist, electrician, officer, pathologist, teacher, lawyer, planner, practitioner, plumber, instructor, surgeon, veterinarian, paramedic, examiner, chemist, machinist, architect, hairdresser, baker, programmer, scientist, dispatcher, cashier, auditor, painter, broker, chef, doctor, firefighter, secretary \\
\bottomrule
\end{tabular}%
\caption{Word-lists considered for finding the \emph{gender subspace} in the input embeddings of the language model. This subset of gendered words is also used for finding the PMI associations. The occupation words are the same as in Table \ref{tab:labour_stats}.}
\label{tab:wordlists}
\end{table*}

\begin{table*}
    \centering
    \begin{tabular}[t]{cc}
    \toprule
    Male & Female \\
    \midrule
        \textbf{man} & \textbf{woman} \\
        \textbf{boy} & \textbf{girl} \\
        guy & gal\\
        gentleman & lady\\
        lord & lady\\
        Mister & Miss\\
        Mr. & Ms.\\
        Mr. & Mrs.\\
        \textbf{male} & \textbf{female}\\
        masculine & feminine\\
        \bottomrule
    \end{tabular}
        \begin{tabular}[t]{cc}
    \toprule
    Male & Female \\
    \midrule
        king & queen\\
        prince & princess\\
        Baron & Baroness\\
        duke & duchess\\
        monk & nun\\
        wizard & witch\\
        landlord & landlady\\
        \bottomrule
    \end{tabular}
    \begin{tabular}[t]{cc}
    \toprule
    Male & Female \\
    \midrule
        \textbf{he} & \textbf{she} \\
        him & her \\
        \textbf{himself} & \textbf{herself} \\
        \textbf{his} & \textbf{her}\\
        his & hers\\
        \bottomrule
    \end{tabular}
    \begin{tabular}[t]{cc}
    \toprule
    Male & Female \\
    \midrule
        \textbf{father} & \textbf{mother}\\
        dad & mum\\
        brother & sister\\
        nephew & niece\\
        uncle & aunt\\
        grandfather & grandmother\\
        \textbf{son} & \textbf{daughter}\\
        grandson & granddaughter\\
        son-in-law & daughter-in-law\\
        stepfather & stepmother\\
        stepson & stepdaughter\\
        father-in-law & mother-in-law\\
        bridegroom & bride\\
        groom & bride\\
        husband & wife\\
        boyfriend & girlfriend\\
        godfather & godmother\\
        \bottomrule
    \end{tabular}\\[5pt]
    \caption{List of gendered word-pairs used in Section \ref{sec:experiments-gender-representation}.
    The word-pairs in bold are also used for finding the gender subspace and computing the PMI associations in Section \ref{sec:experiments-characterizing}.
    We enrich this list by also incorporating the capitalised and pluralised versions of the pairs that are present in the model vocabulary.
    }
    \label{tab:gendered_wordlists}
\end{table*}

\section{Extra figures}
\begin{figure*}[htbp]
    \centering
    \begin{subfigure}[b]{0.7\textwidth}
    \includegraphics[width=\textwidth]{figures/gender_classication.pdf}
    \caption{}\label{fig:gender_classification_appendix}
    \end{subfigure}
    \hfill
    \begin{subfigure}[b]{0.29\textwidth}
    \centering
    \includegraphics[width=\textwidth]{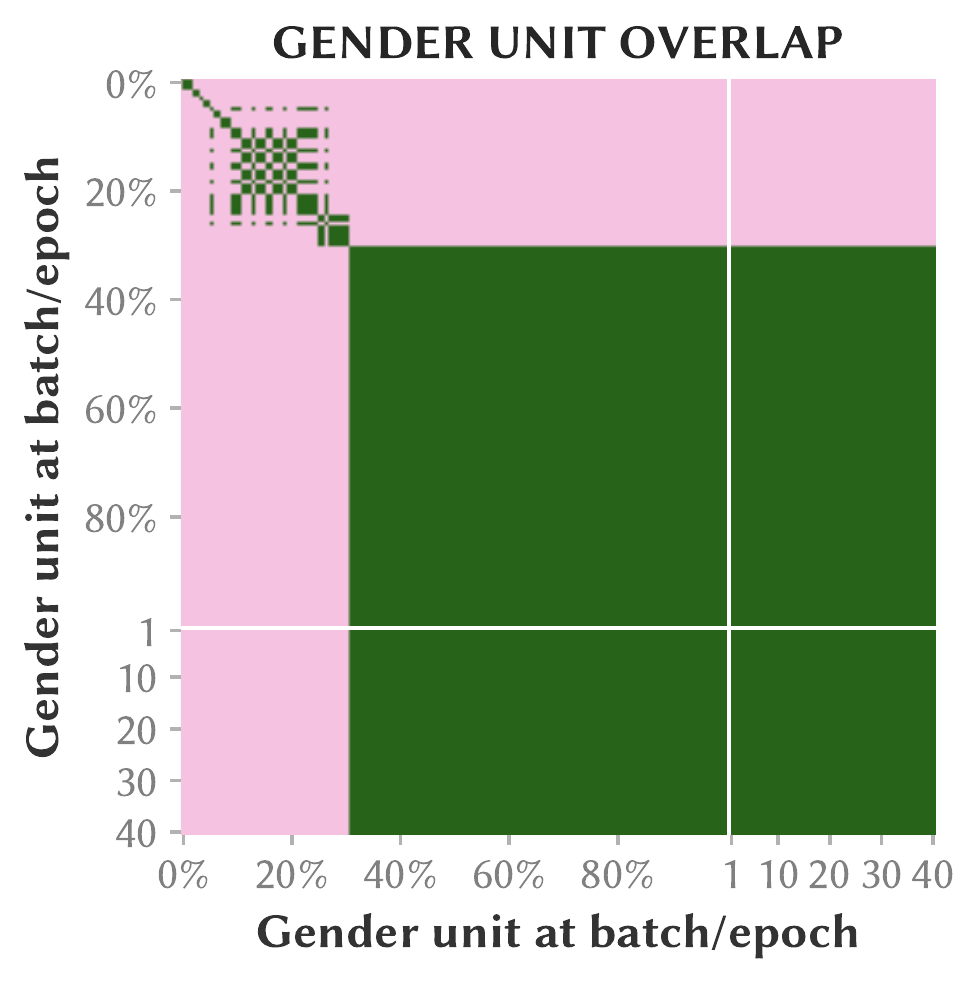}
    \caption{}\label{fig:axis_overlap}
    \end{subfigure}
    
    \caption{Classification accuracy of gender using three different classifiers, that use only the dominant gender unit (green), all other units (red), or all units (orange). Curves show results over training time, averaged across seeds.
    Gender unit overlap (\textit{right}) shows the equality of the principal gender units across time, with green indicating units being equal.
    }
    \label{fig:local_gender_appendix}
\end{figure*}

\begin{figure*}
    \centering
    \includegraphics[width=\textwidth,trim=0cm 9.5cm 0cm 0cm, clip]{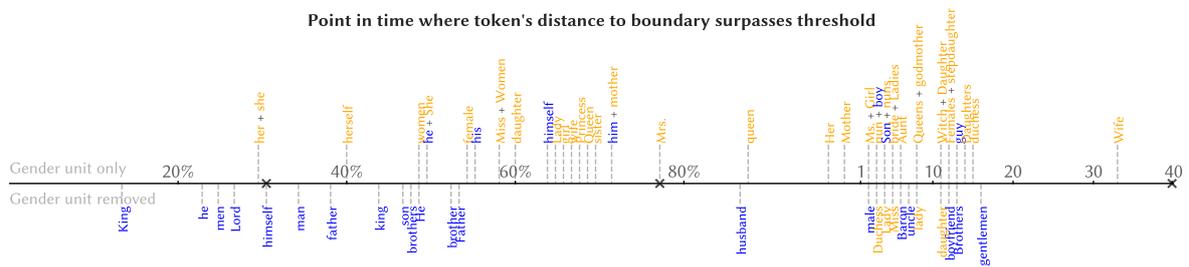}
    \caption{Point in time where a token's distance to the gender decision boundary surpassed a threshold.
    Tokens at the top are based on the single gender unit classifiers; tokens at the bottom are bottom are based on the classifier containing all but the gender unit.
    }
    \label{fig:threshold_tokens}
\end{figure*}

\begin{figure*}[htbp]
    \centering
    \includegraphics[width=\textwidth,trim=0cm 0cm 0cm 0cm, clip]{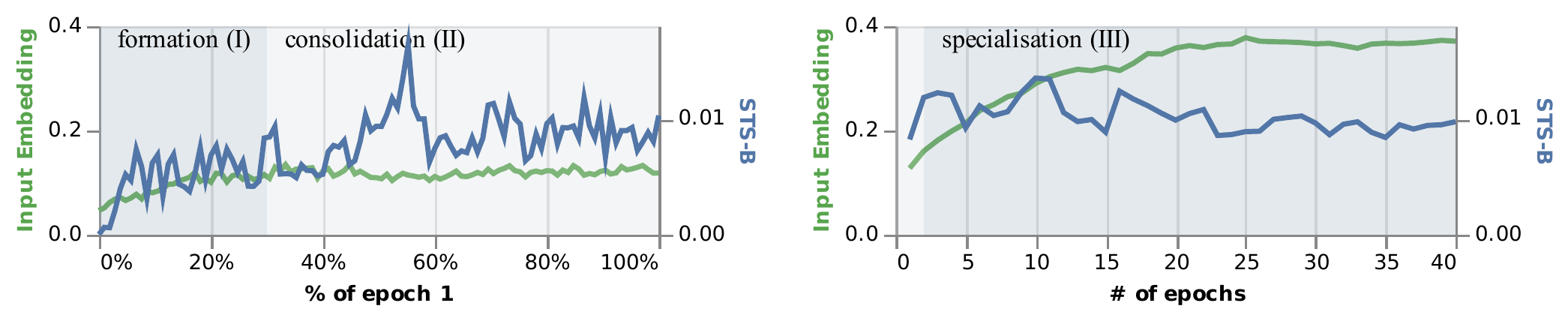}
    \caption{Average absolute bias scores over the occupation terms for the input embeddings and downstream STS-B task.}
    \label{fig:avg_bias}
\end{figure*}

Figures \ref{fig:local_gender_appendix} and \ref{fig:threshold_tokens} support Section \ref{sec:experiments-gender-representation}, while we refer to Figure \ref{fig:avg_bias} in Section \ref{sec:experiments-characterizing}.

